\documentclass[11pt]{article}

\usepackage[final]{acl}

\usepackage{times}
\usepackage{latexsym}

\usepackage[T1]{fontenc}

\usepackage[utf8]{inputenc}

\usepackage{microtype}

\usepackage{inconsolata}

\usepackage{graphicx}
\usepackage{url}
\usepackage{authblk}
\usepackage{setspace}
\usepackage{hyperref}
\usepackage{caption}
\usepackage{booktabs}
\usepackage{tabularx}
\usepackage[normalem]{ulem}
\usepackage{xcolor}
\usepackage{multirow}

\title{A Context-Aware Dataset for Stance Detection in Bioethical Controversies on Reddit}
\author{
\textbf{Hu Huang}\textsuperscript{1},
\textbf{Genan Dai}\textsuperscript{2}, 
\textbf{Fuqiang Niu}\textsuperscript{1},
\textbf{Yi Yang}\textsuperscript{2}, 
\textbf{Zhaoya Gong}\textsuperscript{3},  
\textbf{Bowen Zhang}\textsuperscript{2}\\
  \textsuperscript{1}School of Cyber Science and Technology, \\University of Science and Technology of China, Hefei, China\\
  \textsuperscript{2}School of Artificial Intelligence, Shenzhen Technology University, Shenzhen, China\\
  \textsuperscript{3}School of Urban Planning and Design, Peking University, Shenzhen, China\\
  
  }

\begin{document}
\maketitle

\begin{abstract}
Bioethical debates increasingly unfold on social media, yet stance detection research lacks large-scale, domain-specific resources for modeling such context-dependent discourse. We present BioStance, a context-aware dataset of 39,600 annotated Post-Comment pairs from Reddit bioethical discussions. BioStance covers six controversial targets across three dimensions of bioethical controversy: fundamental value conflicts, individual liberty versus collective responsibility, and technological uncertainty. Each instance preserves hierarchical conversational context and is labeled by three independent annotators using a three-class stance scheme: Favor, Against, and None. The annotations achieve a mean Krippendorff's $\alpha$ of 0.82, indicating substantial reliability. By combining thematic diversity, conversational structure, and high-quality human annotation, BioStance supports research on context-aware stance detection, argument mining, and computational analysis of bioethical discourse.
\end{abstract}

\section{Introduction}

The rapid evolution of biotechnology has reshaped public discourse, moving bioethical debates from academic settings to social media platforms such as \textit{Reddit}\footnote{https://www.reddit.com/}~\cite{scheufele2019science, proferes2021studying}. 
These online discussions provide valuable evidence for understanding how the public negotiates moral values, scientific uncertainty, and socially contested biomedical issues. 
Such analysis is important for computational social science and the development of value-aligned Artificial Intelligence~\cite{weidinger2022taxonomy, 10.1145/3442188.3445922}. 
However, existing stance detection and argument mining datasets are often limited to narrow political topics or event-specific debates, such as elections and COVID-19~\cite{mohammad2016semeval, glandt2021stance, zhang2025logic}, and therefore do not sufficiently cover the broader landscape of biomedical ethics~\cite{cabrio2018five}. 
Moreover, many existing corpora are based on short, context-deprived texts, such as microblogs, which makes it difficult to model the conversational context required for interpreting complex moral reasoning~\cite{habernal2017argumentation, niu2024multimodal}. 
There remains a lack of large-scale, domain-specific, and context-aware datasets for stance detection in bioethical discourse.
Recent model-level advances, such as logic-augmented multi-decision fusion for social-media stance detection, further highlight the need for datasets that expose contextual and target-specific reasoning challenges \citep{zhang2025logic}.

To address this gap, we present \textbf{BioStance} (\textbf{Bio}Ethical \textbf{Stance} Detection Dataset), a large-scale context-aware dataset for stance detection and argument analysis in bioethical discussions. 
BioStance contains 39,600 annotated ``Post-Comment'' instances collected from Reddit, where each comment is paired with its original source post to preserve the discursive context in which stance is expressed.

BioStance covers six controversial bioethical targets organized into three theoretical dimensions of public debate. 
The first dimension, fundamental value conflicts, includes ``Abortion'' and ``Euthanasia'', which involve enduring disputes over individual rights, moral status, and deontological reasoning~\cite{gillon2003ethics}. 
The second dimension, individual liberty versus collective responsibility, includes ``Mandatory Vaccination'' and ``Organ Transplant'', where discussions center on personal autonomy, public welfare, social responsibility, and the allocation of shared medical resources~\cite{kass2001ethics}. 
The third dimension, technological uncertainty and risk, includes ``Gene Editing'' and ``Human Clinical Trials'', which capture public reasoning about scientific trust, experimental risk, and the uncertain consequences of emerging biomedical technologies~\cite{ormond2017human, london2020against}.

A key feature of BioStance is its preservation of the hierarchical ``Post-Comment'' structure. 
By retaining both the source post and the user response, the dataset provides semantic and pragmatic context for modeling stance expressions that may be implicit, ambiguous, or dependent on prior discourse~\cite{habernal2017argumentation, zhang2024knowledge}. 
To ensure annotation quality, each instance was independently labeled by three qualified annotators using a three-class scheme, and the dataset achieves a mean Krippendorff's $\alpha$ above 0.80~\cite{hayes2007answering}.

In summary, BioStance contributes 
(1) {thematic diversity} across established ethical debates, policy-driven public bioethical issues, and emerging biotechnological controversies; 
(2) {contextual depth} through preserved Post-Comment pairs and hierarchical conversation structure; and 
(3) {high-quality human annotations} for training and evaluating context-aware stance detection systems. 
BioStance thus supports research in computational social science, argument mining, biomedical ethics, and bioethical-focused natural language processing.

\section{Methods}

\paragraph{Data Collection.}

We constructed BioStance from Reddit discussions, whose threaded structure makes it suitable for modeling context-dependent stance expression in controversial bioethical debates~\cite{proferes2021studying}. 
The dataset focuses on six bioethical targets: ``Abortion''~\cite{bearak2020unintended}, ``Euthanasia''~\cite{emanuel2016attitudes}, ``Gene Editing''~\cite{baltimore2015prudent}, ``Human Clinical Trials''~\cite{10.1001/jama.283.20.2701}, ``Mandatory Vaccination''~\cite{omer2019mandate}, and ``Organ Transplant''~\cite{persad2009principles}. 
These targets were selected to cover diverse ethical, legal, and scientific dimensions of contemporary bioethical discourse.

We collected Reddit posts and comments through the official API over the period from February 4, 2009, to May 26, 2025. 
For each target, we designed target-specific query lexicons and iteratively refined them to balance topical coverage and retrieval precision. 
The complete query lexicons are provided in Appendix~\ref{app:keywords}, Table~\ref{tab:keywords}. 
The initial crawling process yielded approximately 958,760 candidate threads. 
We treated the ``Post-Comment'' pair as the basic unit of BioStance, where the source post provides context and the comment represents the stance-bearing response. 
Additional details about data sources, subreddit selection, and the construction pipeline are provided in Appendix~\ref{app:construction}.

\paragraph{Preprocessing.}

The raw Reddit data contained deleted content, irrelevant discussions, duplicated posts, and non-standard formatting. 
We therefore applied a multi-stage preprocessing pipeline, including text cleaning, anonymization, relevance filtering, de-duplication, and structural formatting. 
Posts and comments marked as ``[deleted]'' or ``[removed]'' were discarded, and user-identifying information was removed to protect privacy. 
We also filtered out extremely short comments to ensure sufficient semantic content for stance analysis~\cite{niu-etal-2024-challenge}.

Because keyword-based retrieval may introduce false positives, we used GPT-3.5 as an auxiliary binary relevance filter to determine whether each candidate post was substantively related to its intended bioethical target~\cite{gilardi2023chatgpt}. 
The model was used only for topical relevance filtering and did not participate in stance labeling, annotation guidance, or final label decisions. 
After preprocessing, relevance filtering, and de-duplication, we obtained the final dataset for annotation. 
Summary statistics are shown in Table~\ref{tab:data_stats}; detailed preprocessing rules are provided in Appendix~\ref{app:preprocessing}.

\begin{table}[htbp]
    \centering
    \small
    \begin{tabular}{l c c}
        \toprule
        \textbf{Target} & \textbf{\# Samples} & \textbf{Avg. WC} \\
        \midrule
        Abortion & 7,135 & 41.0 \\
        Euthanasia & 7,786 & 47.1 \\
        Gene Editing & 5,712 & 42.3 \\
        Human Clinical Trials & 4,900 & 42.6 \\
        Mandatory Vaccination & 7,673 & 40.1 \\
        Organ Transplant & 6,394 & 41.0 \\
        \midrule
        \textbf{Total} & \textbf{39,600} & \textbf{42.4} \\
        \bottomrule
    \end{tabular}
    \caption{\label{tab:data_stats} Statistics of the preprocessed BioStance. Sample counts are from the final annotated set, and Avg. WC denotes the average word count per sample.}
   
\end{table}

\paragraph{Data Annotation.}
\label{sec:quality_stats}

Following prior stance detection research~\cite{mohammad2016semeval}, we adopted a three-class annotation scheme: Favor, Against, and None. 
Each ``Post-Comment'' instance was independently labeled by three qualified annotators, and the final gold label was determined by majority vote~\cite{dawid1979maximum}. 
To assess annotation quality, we computed inter-annotator agreement after annotation completion~\cite{artstein2008inter}. 
The detailed reliability results are reported in the Technical Validation section. 
The dataset file organization, field definitions, and schema-level details are provided in Appendix~\ref{app:data_records}.

\section{Technical Validation}
This section provides a technical validation of the BioStance dataset, focusing on annotation reliability, label distribution, conversational structure, and topic coverage. The analyses presented here aim to assess the internal consistency, structural integrity, and plausibility of the dataset rather than task performance.

\paragraph{Annotation Reliability.}
To evaluate the reliability of stance annotations, we measured inter-annotator agreement using Krippendorff’s $\alpha$~\cite{hayes2007answering} and overall agreement rates. Each Post-Comment instance was independently annotated by three annotators, as described in the Methods section.

As summarized in Table~\ref{tab:agreement}, the dataset demonstrates substantial agreement across all six bioethical targets. Krippendorff’s $\alpha$ values range from 0.74 to 0.96, with an average $\alpha$ of 0.82. Overall agreement rates range from 86.71\% to 95.15\%, with a mean agreement of 91.33\%. These values indicate a high level of consistency among annotators for a subjective task such as stance detection~\cite{artstein2008inter}.

Targets grounded in scientific or procedural contexts, such as Human Clinical Trials and Gene Editing, exhibit particularly high agreement, suggesting clearer stance articulation in discussions involving technical or institutional considerations. Topics involving fundamental moral conflicts, such as Abortion and Euthanasia, show slightly lower but still substantial agreement, consistent with the inherently subjective nature of these debates~\cite{walker2012corpus}.

To further understand sources of annotation disagreement, we conducted a qualitative analysis of low-consistency instances. Disagreements primarily arise from implicit stance expressions, including irony, sarcasm, and rhetorical exaggeration, where stance is conveyed indirectly through tone or contextual cues rather than explicit evaluative language~\cite{somasundaran2010recognizing}. Additional ambiguity stems from indirect references to the target, topic shifts within multi-turn discussions, and the use of weak-polarity expressions (e.g., “maybe,” “should”) or informal slang. These challenging cases were retained to reflect the complexity of real-world bioethical discourse.

\begin{table}[h]
\centering

\resizebox{\linewidth}{!}{%
\begin{tabular}{l c c}
\hline
\textbf{Target} & {Krippendorff’s $\alpha$} & \textbf{Agreement (\%)} \\ \hline
Abortion & 0.79 & 92.63 \\
Euthanasia & 0.76 & 87.50 \\
Gene Editing & 0.83 & 91.17 \\
Human Clinical Trials & 0.96 & 95.15 \\
Mandatory Vaccination & 0.74 & 86.71 \\
Organ Transplant & 0.85 & 93.73 \\ \hline
\textbf{Average} & \textbf{0.82} & \textbf{91.33} \\ \hline
\end{tabular}
}
\caption{Annotation consistency (Krippendorff’s $\alpha$) and Agreement scores per target.}
\label{tab:agreement}
\end{table}

\paragraph{Dataset Size and Label Distribution.}
The final BioStance dataset contains 39,600 annotated instances across six bioethical targets. This scale is substantial compared to existing stance detection benchmarks~\cite{mohammad2016semeval} and supports robust empirical analysis across diverse bioethical domains.

Table~\ref{tab:label_dist} presents the distribution of stance labels (Favor, Against, None) for each target. The dataset exhibits natural class imbalance, reflecting patterns observed in real-world discussions rather than artifacts introduced during data construction~\cite{10.1145/3369026}. For example, discussions on Human Clinical Trials are predominantly supportive (62.5\% Favor), whereas Mandatory Vaccination elicits relatively higher opposition (41.7\% Against). Other targets, such as Organ Transplant, show a higher proportion of neutral or descriptive stances.

Importantly, no target is dominated by a single stance category, and all three labels are represented across all topics. This distribution suggests that the dataset captures a diverse range of viewpoints while avoiding extreme skew that could undermine its usability for stance detection research.

\begin{table*}[h]
\centering
\small
\begin{tabular}{l|rr|rr|rr|r}
\hline
\textbf{Target} & \textbf{Against} & \textbf{\%} & \textbf{Favor} & \textbf{\%} & \textbf{None} & \textbf{\%} & \textbf{Total} \\ \hline
Abortion & 2,542 & 35.6 & 3,456 & 48.4 & 1,137 & 15.9 & 7,135 \\
Euthanasia & 1,938 & 24.9 & 4,214 & 54.1 & 1,634 & 21.0 & 7,786 \\
Gene Editing & 1,185 & 20.7 & 2,153 & 37.7 & 2,374 & 41.6 & 5,712 \\
Human Clinical Trials & 1,239 & 25.3 & 3,062 & 62.5 & 599 & 12.2 & 4,900 \\
Mandatory Vaccination & 3,200 & 41.7 & 2,558 & 33.3 & 1,915 & 25.0 & 7,673 \\
Organ Transplant & 524 & 8.2 & 2,121 & 33.2 & 3,749 & 58.6 & 6,394 \\ \hline
\textbf{Total} & \textbf{10,628} & \textbf{26.8} & \textbf{17,564} & \textbf{44.4} & \textbf{11,408} & \textbf{28.8} & \textbf{39,600} \\ \hline
\end{tabular}%
\caption{Label distribution of the BioStance. }
\label{tab:label_dist}
\end{table*}

\paragraph{Conversation Structure and Depth.}
Unlike flat sentence-level datasets, BioStance preserves the hierarchical structure of Reddit discussions. Each instance is associated with a conversation depth indicating its position within the discussion tree, enabling analysis of stance expression in both immediate reactions and extended exchanges.

Table~\ref{tab:depth_stats} summarizes the distribution of instances by conversation depth. Top-level posts (Depth 1) account for 5.5\% of the dataset, while the majority of instances (94.5\%) are comments. Direct replies (Depth 2) constitute 61.0\% of the data, and a substantial portion of instances (approximately 33\%) occur at deeper levels (Depth 3 and above). 
This distribution confirms that the dataset includes both shallow and deeply nested interactions.

Average word counts remain relatively stable across depth levels, indicating structural consistency and the absence of truncation or formatting artifacts during preprocessing. The presence of deeper conversational context supports the development and evaluation of context-aware stance detection models that rely on discourse-level information~\cite{sun2018stance}.

\begin{table}[h]
\centering
\small
\begin{tabular}{l c c r}
\hline
\textbf{Instance} & \textbf{Avg. WC} & \textbf{Depth} & \textbf{Number (\%)} \\ \hline
Post & 19.82 & 1 & 2,177 (5.5\%) \\ \hline
\multirow{5}{*}{Comment} & 44.17 & 2 & 24,137 (61.0\%) \\
 & 42.64 & 3 & 7,422 (18.7\%) \\
 & 42.87 & 4 & 3,910 (9.9\%) \\
 & 44.24 & 5 & 1,465 (3.7\%) \\
 & 44.48 & 6 & 489 (1.2\%) \\ \hline
\end{tabular}
\caption{Statistics of the BioStance dataset by conversation depth. Avg. WC denotes the average word count per sample, computed on the preprocessed text field.}
\label{tab:depth_stats}
\end{table}

\paragraph{Topic Coverage and Sanity Checks.}

To assess topical coverage and ensure the absence of systematic retrieval bias, we examined the distribution of sub-topics associated with each target. Table~\ref{tab:keywords} summarizes representative sub-topics and keywords covered within each bioethical domain.

The results indicate that each target encompasses a diverse set of related themes rather than a narrow subset driven by specific keywords. For example, Abortion-related discussions include legal, ethical, and access-oriented perspectives, while Gene Editing spans both technical mechanisms and ethical risk considerations. This diversity suggests that the keyword-based retrieval strategy did not overly constrain topical coverage.

Additional sanity checks confirm that all targets contain sufficient samples across stance categories and conversation depths. No anomalous patterns, such as missing labels or structurally incomplete threads, were observed. Together, these analyses support the internal coherence, coverage adequacy, and structural validity of the BioStance dataset~\cite{gebru2021datasheets}.




Additional qualitative validation results, including word-cloud visualizations, are provided in Appendix~\ref{app:wordcloud}.
Usage guidelines for BioStance, including target-specific and multi-target settings, contextual reconstruction, and label interpretation, are provided in Appendix~\ref{app:usage}.

\section{Potential Research Applications}

BioStance supports several research settings beyond standard supervised stance classification. 
Its target-level organization enables both target-specific evaluation and cross-target transfer analysis, allowing researchers to examine whether models rely on topic-specific lexical cues or learn more general patterns of moral and argumentative reasoning. 
Its preserved Post-Comment structure also makes the dataset suitable for context-aware stance modeling, where models can be compared under different input settings, such as comment-only, post-comment, or broader conversational-context configurations. 
In addition, the six bioethical targets cover distinct forms of public controversy, including enduring moral conflicts, policy-driven collective-responsibility issues, and emerging technological risks. 
BioStance can therefore serve both as an NLP benchmark and as an empirical resource for computational social science research on online bioethical discourse.

\section{Conclusion}

We presented BioStance, a context-aware dataset for stance detection in bioethical controversy. 
BioStance contains 39,600 annotated Post-Comment pairs from Reddit, covering six bioethical targets across three dimensions of public debate. 
By preserving hierarchical conversational context and providing reliable human annotations, the dataset supports the analysis of stance expression in morally complex and context-dependent online discussions. 
Technical validation shows substantial annotation reliability, diverse label distributions, and meaningful conversational depth across targets. 
We expect BioStance to facilitate future research on context-aware stance detection, argument mining, computational social science, and bioethical natural language processing.

\section*{Limitations}

Several limitations of the BioStance dataset should be considered when interpreting results derived from it.
First, all data are sourced from Reddit, which represents a specific online community with platform-dependent discourse norms and demographic biases. As a result, the dataset may not fully reflect opinions or language use in other social media platforms or offline contexts.
Second, the dataset is restricted to English-language content and a single data source. Cross-lingual and cross-platform generalization is therefore beyond the scope of the current release.
Third, stance annotations are explicitly defined with respect to predefined bioethical targets. Instances should not be repurposed for general sentiment analysis or target-agnostic stance classification without careful reinterpretation of the labels.
Finally, although annotation reliability is high, stance expression in bioethical discussions is often implicit, ironic, or context-dependent. Ambiguity is particularly prominent in the none category, which may introduce uncertainty for fine-grained classification tasks.



\section*{Ethical Considerations}

This study uses publicly available data from Reddit. 
All data were collected in accordance with Reddit's terms of service and API usage policies. 
The research involved no interaction with Reddit users and no intervention in online communities. 
To minimize potential privacy risks, all user identifiers were removed, and no attempts were made to re-identify individuals. 
Usernames, user IDs, URLs, and subreddit identifiers were excluded from the released dataset. 
Only anonymized textual content necessary for the analysis was retained and publicly released. 
The dataset is intended for research purposes and should not be used to identify or profile individual users or communities.

The stance labels were produced by human annotators. 
Annotators were given label definitions and instructed to assign one of three stance labels, Favor, Against, or None, with respect to the predefined bioethical target. 
They were instructed to consider the source post as conversational context and to label the stance expressed by the comment, rather than general sentiment. 
The annotation guidelines are provided in Appendix~\ref{app:annotation_guidelines}. 
Annotators were recruited through a crowdsourcing platform and were required to satisfy qualification criteria for English text annotation tasks. 
They were compensated according to the platform's payment policy.

According to applicable institutional guidelines, the use of publicly available and anonymized Reddit data without user interaction does not constitute human-subjects research, and therefore institutional review board (IRB) approval was not required.

\bibliography{custom}

\appendix

\section{Additional Data Construction Details}
\label{app:construction}

\begin{figure}[htbp]
    \centering
    \includegraphics[width=\linewidth]{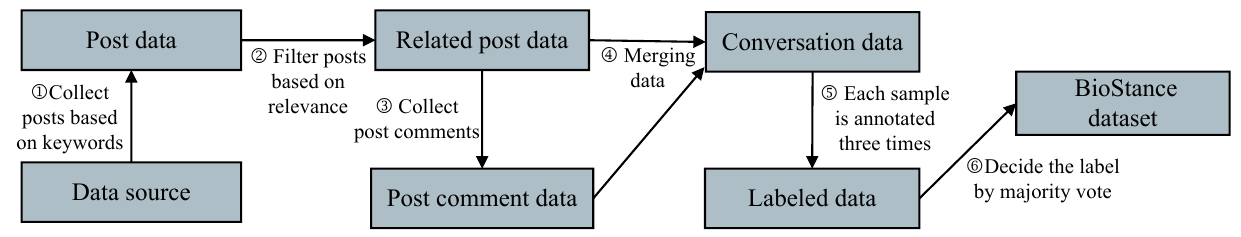}
    \caption{BioStance dataset construction process. The pipeline includes collecting posts, filtering, collecting comments, merging data, annotation, and majority voting.}
    \label{fig:construction_process}
\end{figure}

This appendix provides additional details about the BioStance data construction process. 
The overall pipeline is illustrated in Figure~\ref{fig:construction_process}. It includes collecting Reddit posts based on target-specific keywords, filtering posts for topical relevance, collecting post comments, merging posts and comments into conversation structures, annotating each sample three times, and assigning the final label by majority vote.

During data collection, we targeted subreddits known for in-depth discussions of biomedical, scientific, and ethical issues, including \textit{r/abortion}, \textit{r/science}, \textit{r/biology}, \textit{r/clinicalresearch}, \textit{r/Health}, and \textit{r/medicine}. 
For each target, we collected source posts, associated popularity metrics, and the full hierarchy of user comments. 
The resulting raw corpus contained approximately 958,760 candidate threads before preprocessing and relevance filtering.

\section{Preprocessing Details}
\label{app:preprocessing}

This appendix describes the detailed preprocessing rules used to construct the final BioStance dataset.

First, we removed posts and comments marked as ``[deleted]'' or ``[removed]'' by Reddit or community moderators. 
We then applied standard text normalization, including UTF-8 conversion, removal of non-printable control characters and malformed Unicode artifacts, and removal of URLs, HTML tags, and excessive whitespace~\cite{manning2014stanford}. 
Emojis and common symbols were preserved when they contributed to the meaning of informal social media discourse.

To protect user privacy, we removed user mentions and personally identifiable metadata. 
No private or non-public user metadata was retained~\cite{cassell2000principles}. 
We also discarded comments with fewer than fifteen words to ensure that each instance contained sufficient semantic content for stance detection~\cite{niu-etal-2024-challenge}.

For topical relevance filtering, GPT-3.5 was used only as a binary classifier to determine whether a candidate post was substantively related to the intended bioethical target. 
It was not used for stance annotation, annotation instruction, label adjudication, or any decision regarding Favor, Against, or None labels. 
Within the relevant set, we removed entries with mismatched image-text content and applied de-duplication to eliminate identical posts and repeated spam-like content across subreddits~\cite{broniatowski2018weaponized}. 

\section{Target Selection and Query Lexicons}
\label{app:keywords}

This appendix provides the complete target-specific query lexicons used for Reddit data collection.
The six targets were selected to cover distinct dimensions of bioethical controversy, including fundamental value conflicts, individual liberty versus collective responsibility, and technological uncertainty.
Table~\ref{tab:keywords} lists the target-specific keywords and phrases used to retrieve candidate Reddit discussions.

\begin{table*}[ht]
    \centering
    \small
    \begin{tabularx}{\textwidth}{lX}
        \toprule
        \textbf{Target} & \textbf{Keywords and Phrases} \\
        \midrule
        Abortion & Abortion Legalization, Abortion Rights, Pro-Choice, Legal Abortion, Abortion Law, Reproductive Rights, Abortion Access \\
        Euthanasia & Euthanasia, Assisted Suicide, Right to Die, Mercy Killing, End of Life Choice, Physician-Assisted Death, Medical Aid in Dying \\
        Gene Editing & Gene Editing, CRISPR, Genetic Engineering, DNA Editing, Genome Editing, Gene Therapy \\
        Human Clinical Trials & Human Clinical Trials, Medical Trials, Drug Trials, Clinical Research, Human Testing, Experimental Treatment, Medical Experiments \\
        Mandatory Vaccination & Mandatory Vaccination, Vaccine Mandate, Compulsory Vaccination, Required Vaccines, Forced Vaccination, Vaccination Law \\
        Organ Transplant & Organ Transplant, Organ Donation, Organ Harvesting, Transplantation, Donor Organs, Organ Matching, Transplant Surgery \\
        \bottomrule
    \end{tabularx}
    \caption{Target bioethical topics and corresponding keywords used in query lexicons.}
    \label{tab:keywords}
\end{table*}

\begin{table*}[h]
\centering
\small
\begin{tabular}{l p{2.5cm} p{8cm}}
\hline
\textbf{Column Name} & \textbf{Data Type} & \textbf{Description} \\ \hline
\texttt{Id} & \textit{String} & \textit{A unique identifier for each record. It is composed of information such as topic/thread (e.g., ``779-86-2-1") and is used to locate specific comments in the hierarchical structure.} \\
\texttt{Type} & \textit{Categorical} & \textit{`post' indicates a top-level post in a topic; `comment' indicates a reply to a post or comment.} \\
\texttt{ImgId} & \textit{Integer} & \textit{Identifier of the image associated with a post; resolves to a file in the topic's \texttt{image/} folder} \\
\texttt{ParentId} & \textit{String} & \textit{For posts, ParentId is null; for comments, ParentId is the Id of the parent node (either a post or another comment).} \\
\texttt{Topic} & \textit{Categorical} & \textit{The specific bioethical topic discussed (e.g., Abortion, Euthanasia).} \\
\texttt{Text} & \textit{String} & \textit{The pre-processed text content of the post. URLs and user mentions have been removed.} \\
\texttt{Depth} & \textit{Integer} & \textit{The depth of comments within a thread (1 for a post). Used to analyze the discussion structure and the depth of argument development.} \\
\texttt{Children} & \textit{List} & \textit{A list of direct child comment IDs of the current comment (e.g., [`779-86-2-1-1']). An empty list indicates that there are no child comments.} \\
\texttt{Stance} & \textit{Categorical} & \textit{The annotated stance label: `Favor', `Against', or `None'.} \\
\texttt{Date} & \textit{Date} & \textit{The creation time is represented by a UTC second-level timestamp (e.g., 1634648525). This facilitates time series analysis or dialogue process reconstruction.} \\
\texttt{Keywords} & \textit{List} & \textit{Key terms extracted from the text used for initial retrieval.} \\
\hline
\end{tabular}
\caption{Description of the data fields in the dataset files.}
\label{tab:data_records}
\end{table*}

\begin{figure*}[t]
    \centering
    \includegraphics[width=\textwidth]{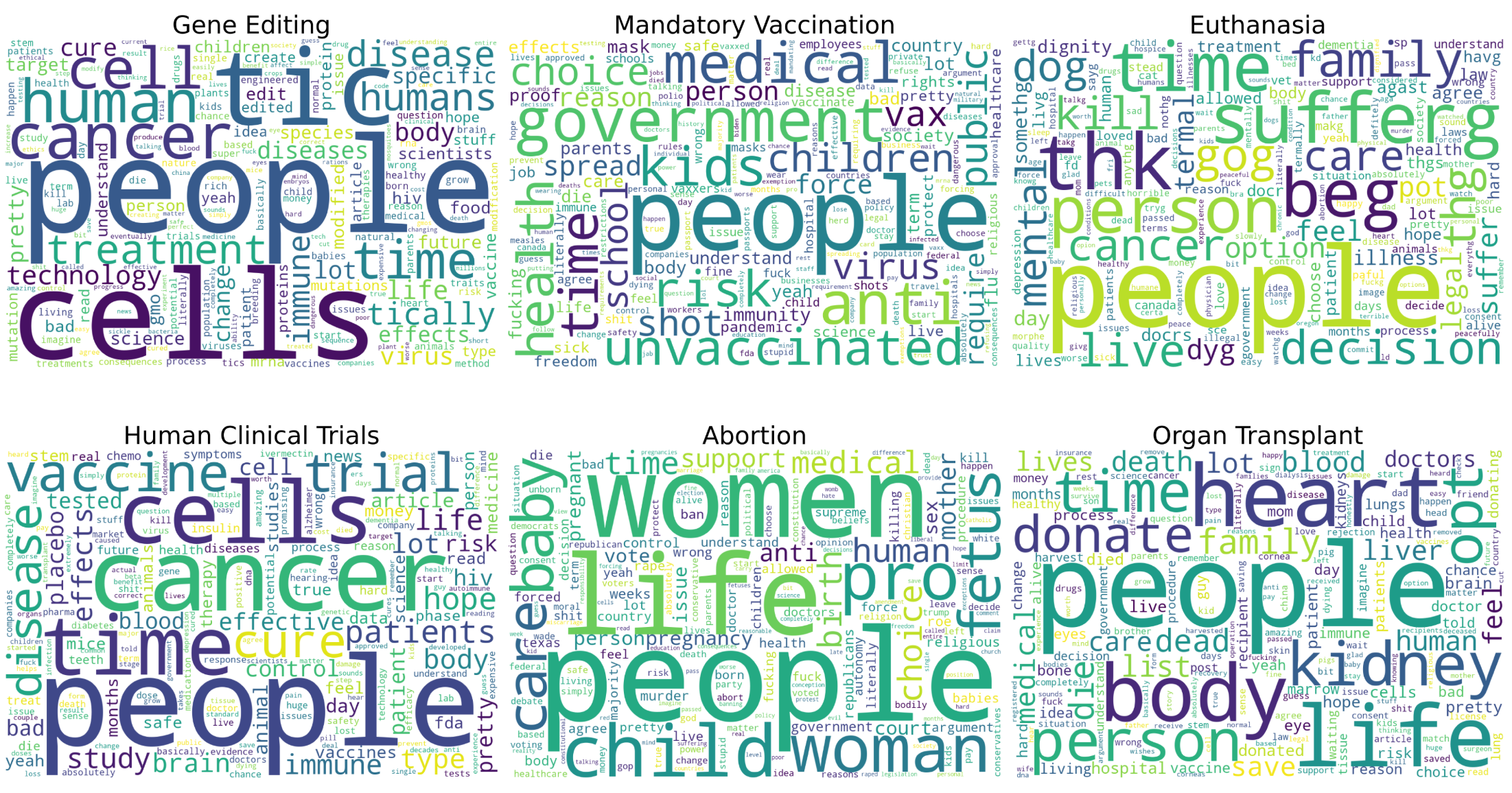}
    \caption{Word cloud image of the data after removing common words.}
    \label{fig:wordcloud}
\end{figure*}

\section{Data Records}
\label{app:data_records}

\paragraph{File Format and Organization.}
{The dataset is organized by topic, where each topic corresponds to a dedicated folder. Each topic folder contains two CSV files (\texttt{posts.csv} and \texttt{comments.csv}) and an optional \texttt{image/} subfolder with images referenced by the posts. All CSV files are UTF-8 encoded to preserve special characters and emojis commonly found in social media text. To protect privacy, the released data does not include direct links to original content, and no user-identifying fields (e.g., usernames or profile IDs) are provided~\cite{williams2017towards}.}

\paragraph{Data Structure.}
{Each topic folder is a self-contained discussion collection. Source posts are provided in \texttt{posts.csv}, while \texttt{comments.csv} contains the associated comments forming reply trees anchored at posts. The thread structure can be reconstructed through parent--child relations, where a comment may reply either to a post or to another comment. Stance labels are provided for both posts and comments.}

Each row in the dataset represents one text instance with its associated metadata and stance label. Table \ref{tab:data_records} summarizes the columns in the provided files.

\section{Qualitative Topic Sanity Check}
\label{app:wordcloud}

This appendix provides a qualitative sanity check of the dominant lexical patterns in BioStance.
Following prior work on word-cloud-based text visualization~\cite{heimerl2014word}, we visualize frequent terms after removing common words to provide an intuitive overview of topic-specific discussion themes.
As shown in Figure~\ref{fig:wordcloud}, the most salient terms are broadly consistent with the corresponding bioethical targets. 
For example, Gene Editing contains terms related to cells, DNA, disease, and treatment; Mandatory Vaccination emphasizes vaccines, government, children, and immunity; and Organ Transplant highlights organ, donor, body, kidney, and medical. 
These patterns suggest that the collected discussions are lexically coherent with their intended targets. 
We use this visualization only as a qualitative sanity check rather than as evidence of annotation quality or model performance.

\section{Usage Notes}
\label{app:usage}

The BioStance dataset is designed to support research on stance expression in bioethical and bioethical-related discussions. This section provides guidance on appropriate usage and interpretation of the data.

The dataset may be used either in a target-specific setting, where instances are restricted to a single bioethical topic, or in a multi-target setting, where data from all topics are merged~\cite{sobhani2017dataset}. Users should note that stance labels are defined with respect to the predefined target associated with each instance, and stance categories are not interchangeable across different targets.

Each record is linked to its conversational context through explicit hierarchical fields. Depending on the research objective, users may treat instances as independent text samples or reconstruct discussion threads to incorporate multi-turn context. When leveraging contextual information, care should be taken to avoid information leakage across experimental splits constructed by users~\cite{glandt2021stance}.

The Date and Keywords fields are provided for traceability and exploratory analysis. The timestamp field enables coarse-grained temporal analysis, while the keyword field reflects the initial retrieval queries and should not be interpreted as ground-truth topical annotations.

The none stance label captures neutral, ambiguous, or target-irrelevant expressions. Users are advised to interpret this category cautiously, as it includes both genuinely neutral content and instances where stance is implicit or context-dependent~\cite{mohammad2016semeval}.
The dataset should not be interpreted as representative of population-level public opinion~\cite{olteanu2019social}.

\section{Annotation Guidelines}
\label{app:annotation_guidelines}

Annotators were asked to assign one of three stance labels to each Post-Comment instance with respect to the predefined bioethical target. 
\textbf{Favor} indicates explicit or implicit support for the target. 
\textbf{Against} indicates explicit or implicit opposition to the target. 
\textbf{None} indicates no clear stance, ambiguity, or content that is not directly relevant to the target. 
Annotators were instructed to consider the source post as context and to label the stance expressed by the comment toward the target, rather than general sentiment or agreement with another user.

\end{document}